\documentclass[11pt]{article}
\usepackage{colacl, epic, eepic, epsfig}
\title{Efficient probabilistic top-down and left-corner parsing${}^{\dag}$}
\author{{\bf Brian Roark and Mark Johnson} \\Cognitive and Linguistic
Sciences\\Box 1978, Brown University\\Providence, RI  02912, USA\\{\tt
brian-roark@brown.edu}\hspace*{.5in}{\tt mj@cs.brown.edu}} 
\begin{document}
\renewcommand{\thefootnote}{\fnsymbol{footnote}}
\maketitle
\begin{abstract}
This paper examines efficient predictive broad-coverage parsing
without dynamic programming. In contrast to bottom-up methods,
depth-first top-down parsing produces partial parses that are fully connected
trees spanning the entire left context, from which any kind of
non-local dependency or partial semantic interpretation can in
principle be read. We contrast two predictive parsing approaches,
top-down and left-corner parsing, and 
find both to be viable. In addition, we find that
enhancement with non-local information not only improves parser
accuracy, but also substantially improves the search efficiency. 
\footnotetext{${}^{\dag}$This material is based on work supported by the National Science Foundation under Grant No. SBR-9720368.} 
\end{abstract}
\bibliographystyle{acl}

\section{Introduction}
\renewcommand{\thefootnote}{\arabic{footnote}}
Strong empirical evidence has been presented over the past 15 years
indicating that the human sentence processing mechanism makes {\it on-line\/}
use of contextual information in the preceding discourse
\cite{Crain85,Altmann88,Britt94} and in the
visual environment \cite{Tanen95}. These results lend
support to Mark Steedman's \shortcite{Steed89} ``intuition'' that sentence
interpretation takes place incrementally, and that partial
interpretations are being built while the sentence is being
perceived. This is a very commonly held view among psycholinguists
today. 

Many possible models of human sentence processing can be made
consistent with the above view, but the general assumption that must
underlie them all is that explicit relationships between lexical items 
in the sentence must be specified incrementally.  Such a processing
mechanism stands in marked contrast to 
dynamic programming parsers, which delay construction of a constituent
until all of its sub-constituents have been completed, and whose
partial parses thus consist of disconnected tree fragments. For
example, such parsers do not integrate a main verb into the same tree
structure as its subject {\small NP} until the {\small VP} has been completely parsed,
and in many cases this is the final step of the entire parsing
process. Without explicit on-line integration, it would be difficult
(though not impossible) to produce partial interpretations
on-line. Similarly, it may be difficult to use non-local statistical
dependencies (e.g. between subject and main verb) to actively guide
such parsers. 

Our predictive parser does not use dynamic programming,
but rather maintains fully connected trees spanning the entire left
context, which make explicit the relationships between constituents
required for partial interpretation. The parser uses probabilistic
best-first parsing methods to pursue the most likely analyses first,
and a beam-search to avoid the non-termination problems typical of
non-statistical top-down predictive parsers. 

There are two main
results. First, this approach works and, with appropriate attention to
specific algorithmic details, is surprisingly efficient. Second, not
just accuracy but also efficiency improves as the language model is
made more accurate. This bodes well for future research into the use
of other non-local (e.g. lexical and semantic) information to guide
the parser.   

In addition, we show that the improvement in accuracy
associated with left-corner parsing over top-down is attributable to
the non-local information supplied by the 
strategy, and can thus be obtained through other methods that utilize
that same information. 

\section{Parser architecture}

The parser proceeds incrementally from left to right, with one item of
look-ahead. Nodes are expanded in a standard top-down, left-to-right
fashion. The parser utilizes: (i) a probabilistic context-free grammar
({\small PCFG}), induced via standard relative frequency estimation from a
corpus of parse trees; and (ii) look-ahead probabilities as described
below. Multiple competing partial parses (or analyses) are held on a
priority queue, which we will call the {\it pending\/} heap. They are ranked
by a figure of merit ({\small FOM}), which will be discussed below. Each
analysis has its own stack of nodes to be expanded, as well as a
history, probability, and {\small FOM}. The highest ranked analysis is popped
from the pending heap, and the category at the top of its stack is
expanded. A category is expanded using every rule which could
eventually reach the look-ahead terminal. For every such rule
expansion, a new analysis is created\footnote{We count each of these
as a parser state (or rule expansion) {\it considered\/}, which can be 
used as a measure of efficiency.} and pushed back onto the pending
heap. 

The {\small FOM} for an analysis is the product of the probabilities of
all {\small PCFG} rules used in its derivation and what we call its look-ahead
probability ({\small LAP}). The {\small LAP} approximates the product of the
probabilities of the rules that will be required to link the analysis
in its current state with the look-ahead terminal\footnote{Since this
is a non-lexicalized grammar, we are taking pre-terminal POS markers
as our terminal items.}. That is, for a
grammar {\small G}, a stack state [{\small $C_{1} \dots C_{n}$}] and a
look-ahead terminal item $\omega$: 

\begin{center}(1) $LAP = P_{G}([C_{1} \dots
C_{n}] \stackrel{\star}{\rightarrow} \omega\alpha)$\end{center}

We recursively estimate this with two empirically observed conditional
probabilities for every non-terminal $C_{i}$ on the stack:
$\widehat{P} (C_{i} \stackrel{\star}{\rightarrow} \omega)$
and $\widehat{P} (C_{i} \stackrel{\star}{\rightarrow} \epsilon)$.
The {\small LAP}
approximation for a given stack state and look-ahead terminal is:  

\begin{center}(2) $P_{G}([C_{i} \dots
C_{n}] \stackrel{\star}{\rightarrow} \omega\alpha) $\hspace*{.1in} $
\approx $\hspace*{.1in} $
\widehat{P} (C_{i} \stackrel{\star}{\rightarrow} \omega)$ +\\$
\widehat{P} (C_{i} \stackrel{\star}{\rightarrow} \epsilon) *
P_{G}([C_{i+1} \dots
C_{n}] \stackrel{\star}{\rightarrow} \omega\alpha)$
\end{center}

When the topmost stack category of an analysis matches the look-ahead
terminal, the terminal is popped from the stack and the analysis is
pushed onto a second priority queue, which we will call the {\it success\/}
heap. Once there are ``enough'' analyses on the success heap, all those
remaining on the pending heap are discarded. The success heap then
becomes the pending heap, and the look-ahead is moved forward to the
next item in the input string. When the end of the input string is
reached, the analysis with the highest probability and an empty stack
is returned as the parse. If no such parse is found, an error is
returned.

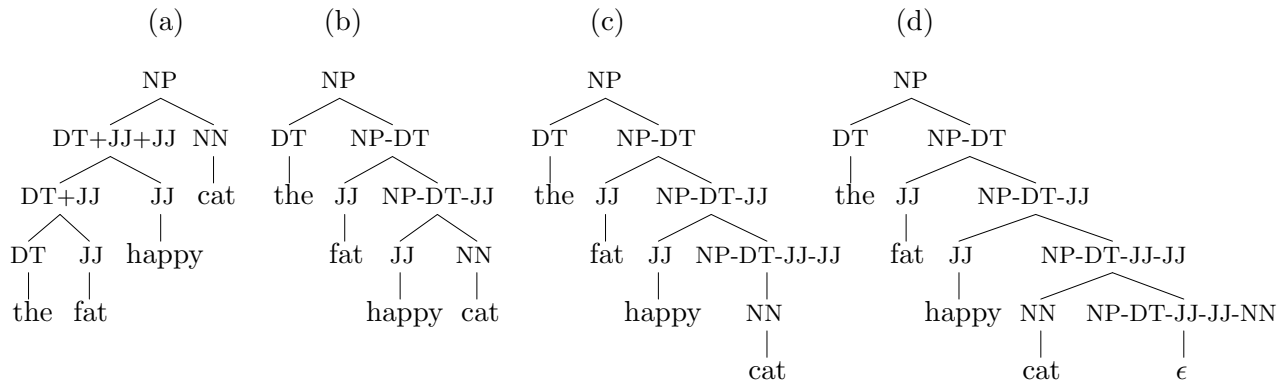
\begin{figure*}
\begin{picture}(95,0)(0,-135)
\put(52,-8){(a)}

\put(50,-30){\footnotesize NP}
\drawline(57,-34)(38,-44)
\put(16,-52){\footnotesize DT+JJ+JJ}
\drawline(38,-56)(19,-66)
\put(4,-74){\footnotesize DT+JJ}
\drawline(19,-78)(7,-88)
\put(-0,-96){\footnotesize DT}
\drawline(7,-100)(7,-110)
\put(1,-118){the}
\drawline(19,-78)(30,-88)
\put(26,-96){\footnotesize JJ}
\drawline(30,-100)(30,-110)
\put(24,-118){fat}
\drawline(38,-56)(57,-66)
\put(53,-74){\footnotesize JJ}
\drawline(57,-78)(57,-88)
\put(44,-96){happy}
\drawline(57,-34)(77,-44)
\put(69,-52){\footnotesize NN}
\drawline(77,-56)(77,-66)
\put(71,-74){cat}
\end{picture}
\begin{picture}(95,0)(0,-135)
\put(20,-8){(b)}

\put(19,-30){\footnotesize NP}
\drawline(26,-34)(7,-44)
\put(-0,-52){\footnotesize DT}
\drawline(7,-56)(7,-66)
\put(1,-74){the}
\drawline(26,-34)(46,-44)
\put(30,-52){\footnotesize NP-DT}
\drawline(46,-56)(28,-66)
\put(24,-74){\footnotesize JJ}
\drawline(28,-78)(28,-88)
\put(22,-96){fat}
\drawline(46,-56)(63,-66)
\put(42,-74){\footnotesize NP-DT-JJ}
\drawline(63,-78)(49,-88)
\put(45,-96){\footnotesize JJ}
\drawline(49,-100)(49,-110)
\put(36,-118){happy}
\drawline(63,-78)(78,-88)
\put(70,-96){\footnotesize NN}
\drawline(78,-100)(78,-110)
\put(72,-118){cat}
\end{picture}
\begin{picture}(110,0)(0,-135)
\put(22,-8){(c)}

\put(21,-30){\footnotesize NP}
\drawline(28,-34)(7,-44)
\put(-0,-52){\footnotesize DT}
\drawline(7,-56)(7,-66)
\put(1,-74){the}
\drawline(28,-34)(48,-44)
\put(32,-52){\footnotesize NP-DT}
\drawline(48,-56)(28,-66)
\put(24,-74){\footnotesize JJ}
\drawline(28,-78)(28,-88)
\put(22,-96){fat}
\drawline(48,-56)(68,-66)
\put(47,-74){\footnotesize NP-DT-JJ}
\drawline(68,-78)(48,-88)
\put(44,-96){\footnotesize JJ}
\drawline(48,-100)(48,-110)
\put(35,-118){happy}
\drawline(68,-78)(89,-88)
\put(62,-96){\footnotesize NP-DT-JJ-JJ}
\drawline(89,-100)(89,-110)
\put(81,-118){\footnotesize NN}
\drawline(89,-122)(89,-132)
\put(82,-140){cat}
\end{picture}
\begin{picture}(150,135)(0,-135)
\put(24,-8){(d)}
\put(23,-30){\footnotesize NP}
\drawline(30,-34)(7,-44)
\put(0,-52){\footnotesize DT}
\drawline(7,-56)(7,-66)
\put(1,-74){the}
\drawline(30,-34)(52,-44)
\put(36,-52){\footnotesize NP-DT}
\drawline(52,-56)(28,-66)
\put(24,-74){\footnotesize JJ}
\drawline(28,-78)(28,-88)
\put(22,-96){fat}
\drawline(52,-56)(77,-66)
\put(55,-74){\footnotesize NP-DT-JJ}
\drawline(77,-78)(48,-88)
\put(44,-96){\footnotesize JJ}
\drawline(48,-100)(48,-110)
\put(35,-118){happy}
\drawline(77,-78)(106,-88)
\put(79,-96){\footnotesize NP-DT-JJ-JJ}
\drawline(106,-100)(79,-110)
\put(71,-118){\footnotesize NN}
\drawline(79,-122)(79,-132)
\put(72,-140){cat}
\drawline(106,-100)(133,-110)
\put(96,-118){\footnotesize NP-DT-JJ-JJ-NN}
\drawline(133,-122)(133,-132)
\put(130,-140){$\epsilon$}
\end{picture}
\caption{Binarized trees:  (a) left binarized ({\small LB}); (b) right
binarized to binary ({\small RB2}); (c) right binarized to unary
({\small RB1}); (d) right binarized to nullary ({\small RB0})}\label{fig:bin}
\end{figure*}

The specifics of the beam-search dictate how many analyses
on the success heap constitute ``enough''. One approach is to set a
constant beam width, e.g. 10,000 analyses on the success heap, at
which point the parser 
moves to the next item in the input. A problem with this approach is
that parses towards the bottom of the success heap may be so unlikely
relative to those at the top that they have little or no chance of
becoming the most likely parse at the end of the day, causing wasted
effort. An alternative approach is to dynamically vary the beam width
by stipulating a factor, say $10^{-5}$, and proceed until the best analysis
on the pending heap has an {\small FOM} less than $10^{-5}$ times the probability of
the best analysis on the success heap. Sometimes, however, the number
of analyses that fall within such a range can be enormous, creating
nearly as large of a processing burden as the first approach. As a
compromise between these two approaches, we stipulated a base beam
factor $\alpha$ (usually $10^{-4}$), and the actual beam factor used
was $\alpha \ast \beta$, where $\beta$ is the number of analyses on
the success heap. Thus, when 
$\beta$ is small, the beam stays relatively wide, to include as many
analyses as possible; but as $\beta$ grows, the beam narrows. We found this
to be a simple and successful compromise. 

Of course, with a left
recursive grammar, such a top-down parser may never terminate. If {\it 
no\/} analysis ever makes it to the success heap, then, however one defines
the beam-search, a top-down depth-first search with a left-recursive
grammar will 
never terminate. To avoid this, one must place an upper bound on the
number of analyses allowed to be pushed onto the pending heap. If that
bound is exceeded, the parse fails. With a left-corner strategy, which
is not prey to left recursion, no such upper bound is necessary. 

\section{Grammar transforms}

\newcite{Nijholt80} characterized parsing strategies in terms of {\it announce
points\/}: the point at which a parent category is announced
(identified) relative to its children, and the point at which the rule
expanding the parent is identified. In 
pure top-down parsing, a parent category and the rule expanding it are
announced {\it before\/} any of its children. In pure bottom-up parsing, they
are identified {\it after\/} all of the children. Grammar transforms are one
method for changing the announce points. In top-down parsing with an
appropriately binarized grammar, the parent is identified {\it before\/}, but
the rule expanding the parent {\it after\/}, all of the children. Left-corner
parsers announce a parent category and its expanding rule {\it after\/} its
leftmost child has been completed, but {\it before\/} any of the other
children. 

\subsection{Delaying rule identification through binarization}
\begin{table*}
\begin{tabular} {|p{.8in}|p{.6in}|p{.65in}|p{.8in}|p{.9in}|p{.7in}|p{.9in}|}
\hline
{\small Binarization} &
{\small Rules in Grammar} &
{\small Percent of Sentences Parsed${}^{\ast}$} &
{\small Avg. States Considered} &
{\small Avg. Labelled Precision and Recall${}^{\dag}$} &
{\small Avg. MLP Labelled Prec/Rec${}^{\dag}$} &
{\small Ratio of Avg. Prob to Avg. MLP Prob${}^{\dag}$} \\\hline
{\small None} &
{\small 14962} &
{\small 34.16} &
{\small 19270} &
{\small .65521} &
{\small .76427} &
{\small .001721} \\\hline
{\small LB} &
{\small 37955} &
{\small 33.99} &
{\small 96813} &
{\small .65539} &
{\small .76095} &
{\small .001440} \\\hline
{\small RB1} &
{\small 29851} &
{\small 91.27} &
{\small 10140} &
{\small .71616} &
{\small .72712} &
{\small .340858} \\\hline
{\small RB0} &
{\small 41084} &
{\small 97.37} &
{\small 13868} &
{\small .73207} &
{\small .72327} &
{\small .443705} \\\hline
\end{tabular}
{\footnotesize Beam Factor = $10^{-4}$ \hspace*{.18in}
${}^{\ast}$Length $\leq$ 40 (2245 sentences
in F23 - Avg. length = 21.68) \hspace*{.18in}
${}^{\dag}$Of those sentences parsed}
\caption{The effect of different approaches to
binarization}\label{tab:bin}
\end{table*}

Suppose that the category on the top of the stack is an {\small $NP$} and there
is a determiner ({\small $DT$}) in the look-ahead. In such a situation, there
is no information to distinguish between the rules \begin{small}$NP
\rightarrow DT$\hspace*{.1in}$JJ$\hspace*{.1in}$NN$\end{small} and
\begin{small}$NP \rightarrow
DT$\hspace*{.1in}$JJ$\hspace*{.1in}$NNS$\end{small}.  If the decision 
can be delayed, however, until such a time as the 
relevant pre-terminal is in the look-ahead, the parser can make a more
informed decision. Grammar binarization is one way to do this, by
allowing the parser to use a rule like \begin{small}$NP \rightarrow
DT$\hspace*{.1in}$NP$-$DT$\end{small}, where the
new non-terminal {\small $NP$-$DT$} can expand into anything that
follows a {\small $DT$}
in an {\small $NP$}. The expansion of {\small $NP$-$DT$} occurs only
after the next pre-terminal is in the look-ahead. Such a delay is
essential for an efficient implementation of the kind of incremental
parser that we are proposing.

There are actually
several ways to make a grammar binary, some of which are better than
others for our parser. The first distinction that can be drawn is
between what we will call {\it left\/} binarization ({\small LB}) versus {\it right\/}
binarization ({\small RB}, see figure \ref{fig:bin}). In the former, the leftmost items
on the righthand-side of each rule are grouped together; in the
latter, the rightmost items on the righthand-side of the rule are
grouped together. Notice that, for a top-down, left-to-right parser,
{\small RB} is the appropriate transform, because it underspecifies the right
siblings. With {\small LB}, a top-down parser must identify all of the
siblings before reaching the leftmost item, which does not aid our
purposes. 

Within {\small RB} transforms, however, there is some variation, with
respect to how long rule underspecification is maintained. One method
is to have the final underspecified category rewrite as a binary rule
(hereafter {\small RB2}, see figure \ref{fig:bin}b). Another is to
have the final underspecified category rewrite as a unary rule
({\small RB1}, figure \ref{fig:bin}c). The last is to have the final
underspecified category rewrite as a nullary rule ({\small RB0},
figure \ref{fig:bin}d). Notice that the original motivation 
for {\small RB}, to delay specification until the relevant items are present
in the look-ahead, is not served by {\small RB2}, because the second child
must be specified without being present in the look-ahead. {\small RB0} pushes
the look-ahead out to the first item in the string {\it after\/} the
constituent being expanded, which can be useful in deciding between
rules of unequal length, e.g. \begin{small}$NP \rightarrow
DT$\hspace*{.1in}$NN$\end{small} and  
\begin{small}$NP \rightarrow
DT$\hspace*{.1in}$NN$\hspace*{.1in}$NN$\end{small}. 

Table \ref{tab:bin} summarizes some trials demonstrating the effect of
different 
binarization approaches on parser performance. The grammars were
induced from sections 2-21 of the Penn Wall St. Journal Treebank
\cite{Marcus93}, and tested on section 23. For each transform
tested, every tree in the training corpus was transformed before
grammar induction, resulting in a transformed {\small PCFG} and look-ahead
probabilities estimated in the standard way. Each parse returned by
the parser was de-transformed for evaluation\footnote{See
\newcite{Johnson98b} for details of the transform/de-transform
paradigm.}. The parser used in each trial was identical, with a base
beam factor $\alpha = 10^{-4}$. The performance 
is evaluated using these measures: (i) the percentage of candidate
sentences for which a parse was found (coverage); (ii) the average
number of states (i.e. rule expansions) considered per candidate
sentence (efficiency); and 
(iii) the average labelled precision and recall of those sentences for
which a parse was found (accuracy). We also used the same grammars
with an exhaustive, bottom-up {\small CKY} parser, to ascertain both the
accuracy and probability of the maximum likelihood parse ({\small MLP}). We
can then additionally compare the parser's performance to the {\small MLP}'s
on those same sentences. 

As expected, {\it left\/} binarization conferred no
benefit to our parser. {\it Right\/} binarization, in contrast, improved
performance across the board. {\small RB0} provided a substantial improvement
in coverage and accuracy over {\small RB1}, with something of a decrease in
efficiency. This efficiency hit is partly attributable to the fact that
the same tree has more nodes with {\small RB0}. Indeed, the efficiency
improvement with right binarization over the standard grammar is even
more interesting in light of the great increase in the size of the
grammars. 

It is worth noting at this point that, with the {\small RB0} grammar,
this parser is now a viable 
broad-coverage statistical parser, with good coverage, accuracy, and
efficiency\footnote{The very efficient bottom-up statistical parser
detailed in \newcite{Charniak98} measured efficiency in terms of total 
edges {\it popped\/}.  An edge (or, in our case, a parser state) is
{\it considered\/} when a probability is calculated for it, and we
felt that this was a better efficiency measure than simply those
popped.  As a baseline, their parser {\it considered\/} an average of
2216 edges per sentence in section 22 of the WSJ corpus (p.c.).}. Next we considered the left-corner parsing strategy.

\subsection{Left-corner parsing}
\begin{table*}
\begin{tabular} {|p{1.05in}|p{.6in}|p{.6in}|p{.75in}|p{.85in}|p{.65in}|p{.85in}|}
\hline
{\small Transform} &
{\small Rules in Grammar} &
{\small Pct. of Sentences Parsed${}^{\ast}$} &
{\small Avg. States Considered} &
{\small Avg Labelled Precision and Recall${}^{\dag}$} &
{\small Avg. MLP Labelled Prec/Rec${}^{\dag}$} &
{\small Ratio of Avg. Prob to Avg. MLP Prob${}^{\dag}$} \\\hline
{\small Left Corner (LC)} &
{\small 21797} &
{\small 91.75} &
{\small 9000} &
{\small .76399} &
{\small .78156} &
{\small .175928} \\\hline
{\small LB $\circ$ LC} &
{\small 53026} &
{\small 96.75} &
{\small 7865} &
{\small .77815} &
{\small .78056} &
{\small .359828} \\\hline
{\small LC $\circ$ RB} &
{\small 53494} &
{\small 96.7} &
{\small 8125} &
{\small .77830} &
{\small .78066} &
{\small .359439} \\\hline
{\small LC $\circ$ RB $\circ$ ANN} &
{\small 55094} &
{\small 96.21} &
{\small 7945} &
{\small .77854} &
{\small .78094} &
{\small .346778} \\\hline
{\small RB $\circ$ LC} &
{\small 86007} &
{\small 93.38} &
{\small 4675} &
{\small .76120} &
{\small .80529} &
{\small .267330} \\\hline
\end{tabular}
{\footnotesize Beam Factor = $10^{-4}$ \hspace*{.18in}
${}^{\ast}$Length $\leq$ 40 (2245 sentences
in F23 - Avg. length = 21.68) \hspace*{.18in}
${}^{\dag}$Of those sentences parsed}
\caption{Left Corner Results}\label{tab:left}
\end{table*}

Left-corner ({\small LC}) parsing \cite{Rosenkrantz70} is a
well-known strategy that uses both bottom-up evidence (from the left
corner of a rule) and top-down prediction (of the rest of the
rule). Rosenkrantz and Lewis showed how to transform a context-free
grammar into a grammar that, when used by a top-down parser, follows
the same search path as an {\small LC} parser. These {\small LC}
grammars allow us to use exactly the same predictive parser to
evaluate top-down versus {\small LC} 
parsing. Naturally, an {\small LC} grammar performs best with our parser when
right binarized, for the same reasons outlined above. We use transform
composition to apply first one transform, then another to the output
of the first. We denote this {\small A} $\circ$ {\small B} where
({\small A} $\circ$ {\small B})(t) = {\small B} ({\small A}
(t)). After applying the left-corner transform, we then binarize the
resulting grammar\footnote{Given that the LC transform involves
nullary productions, the use of RB0 is not needed, i.e. nullary
productions need only be introduced from one source.  Thus
binarization with left corner is always to unary (RB1).}, i.e. {\small LC} $\circ$ {\small RB}. 

Another probabilistic {\small LC} parser investigated \cite{Manning97},
which utilized an {\small LC} parsing architecture (not a transformed
grammar), also got a performance boost through right 
binarization. This, however, is equivalent to {\small RB} $\circ$
{\small LC}, which is a very different grammar from {\small LC}
$\circ$ {\small RB}. Given our two binarization orientations ({\small
LB} and {\small RB}), there are four possible compositions of 
binarization and {\small LC} transforms: 
\begin{center}\begin{small}
(a) LB $\circ$ LC (b) RB $\circ$ LC
(c) LC $\circ$ LB  (d) LC $\circ$ RB 
\end{small}\end{center}
Table \ref{tab:left} shows left-corner results over various
conditions\footnote{Option (c) is not the appropriate kind of
binarization for our parser, as argued in the previous section, and so 
is omitted.}. Interestingly, options (a) and (d) encode the same
information, leading to nearly identical performance\footnote{The
difference is due to the introduction of vacuous unary rules with
RB.}. As stated before, right binarization moves the rule announce
point from before to after all of the children. The {\small LC} transform is
such that {\small LC} $\circ$ {\small RB} 
also delays {\it parent\/} identification until after all of the
children. The transform {\small LC} $\circ$ {\small RB} $\circ$
{\small ANN} moves the parent announce
point back to the left corner by introducing unary rules at the left
corner that simply identify the parent of the binarized rule. This
allows us to test the effect of the position of the parent announce
point on the performance of the parser. As we can see, however, the
effect is slight, with similar performance on all measures. 

{\small RB} $\circ$ {\small LC} performs with higher accuracy than the others when used with
an exhaustive parser, but seems to require a massive beam in order to
even approach performance at the {\small MLP} level. \newcite{Manning97}
used a beam width of 40,000 parses on the success heap at each input
item, which 
must have resulted in an order of magnitude more rule expansions
than what we have been considering up to now, and yet their average
labelled precision and recall (.7875) still fell well below what we
found to be the {\small MLP} accuracy (.7987) for the grammar. We are still
investigating why this grammar functions so poorly when used by an
incremental parser. 

\subsection{Non-local annotation}

\newcite{Johnson98b} discusses the improvement of {\small PCFG} models via the
annotation of non-local information onto non-terminal nodes in the
trees of the training corpus. One simple example is to
copy the parent node onto every non-terminal, e.g. the rule
\begin{small}$S \rightarrow NP$\hspace*{.1in}$VP$\end{small} becomes 
\begin{small}$S \rightarrow
NP^{\uparrow}S$\hspace*{.1in}$VP^{\uparrow}S$\end{small}.  The idea
here is that 
the distribution of rules of expansion of a particular non-terminal
may differ depending on the non-terminal's parent. Indeed, it was
shown that this additional information improves the {\small MLP}
accuracy dramatically.

We looked at two kinds of
non-local information annotation: parent ({\small PA}) and left-corner
({\small LCA}). Left-corner parsing gives improved accuracy over top-down or
bottom-up parsing with the same grammar. Why? One reason may be that
the ancestor category exerts the same kind of non-local influence
upon the parser that the parent category does in parent annotation. To
test this, we annotated the left-corner ancestor category onto every
leftmost non-terminal category. The results of our annotation trials
are shown in table \ref{tab:ann}. 

\begin{table*}
\begin{tabular} {|p{1.05in}|p{.6in}|p{.6in}|p{.75in}|p{.85in}|p{.65in}|p{.85in}|}
\hline
{\small Transform} &
{\small Rules in Grammar} &
{\small Pct. of Sentences Parsed${}^{\ast}$} &
{\small Avg. States Considered} &
{\small Avg Labelled Precision and Recall${}^{\dag}$} &
{\small Avg. MLP Labelled Prec/Rec${}^{\dag}$} &
{\small Ratio of Avg. Prob to Avg. MLP Prob${}^{\dag}$} \\\hline
{\small RB0} &
{\small 41084} &
{\small 97.37} &
{\small 13868} &
{\small .73207} &
{\small .72327} &
{\small .443705} \\\hline
{\small PA $\circ$ RB0} &
{\small 63467} &
{\small 95.19} &
{\small 8596} &
{\small .79188} &
{\small .79759} &
{\small .486995} \\\hline
{\small LC $\circ$ RB} &
{\small 53494} &
{\small 96.7} &
{\small 8125} &
{\small .77830} &
{\small .78066} &
{\small .359439} \\\hline
{\small LCA $\circ$ RB0} &
{\small 58669} &
{\small 96.48} &
{\small 11158} &
{\small .77476} &
{\small .78058} &
{\small .495912} \\\hline
{\small PA $\circ$ LC $\circ$ RB} &
{\small 80245} &
{\small 93.52} &
{\small 4455} &
{\small .81144} &
{\small .81833} &
{\small .484428} \\\hline
\end{tabular}
{\footnotesize Beam Factor = $10^{-4}$ \hspace*{.18in}
${}^{\ast}$Length $\leq$ 40 (2245 sentences
in F23 - Avg. length = 21.68) \hspace*{.18in}
${}^{\dag}$Of those sentences parsed}
\caption{Non-local annotation results}\label{tab:ann}
\end{table*}

There are two important points to notice from
these results. First, with {\small PA} we get not only the previously reported
improvement in accuracy, but additionally a fairly dramatic decrease
in the number of parser states that must be visited to find a
parse. That is, the non-local information not only improves the final
product of the parse, but it guides the parser more quickly to the
final product. The annotated grammar has 1.5 times as many rules, and
would slow a bottom-up {\small CKY} parser proportionally. Yet our parser
actually considers far fewer states en route to the more accurate
parse. 

Second, {\small LC}-annotation gives nearly all of the accuracy gain of
left-corner parsing\footnote{The rest could very well be within
noise.}, in support of the hypothesis that the ancestor 
information was responsible for the observed accuracy
improvement. This result suggests that if we can determine the
information that is being annotated by the troublesome {\small RB} $\circ$ {\small LC}
transform, we may be able to get the accuracy improvement with a
relatively narrow beam. Parent-annotation before the {\small LC} transform gave
us the best performance of all, with very few states considered on
average, and excellent accuracy for a non-lexicalized grammar. 

\section{Accuracy/Efficiency tradeoff}
\begin{figure*}
\hspace*{1.1in}
\epsfig{file=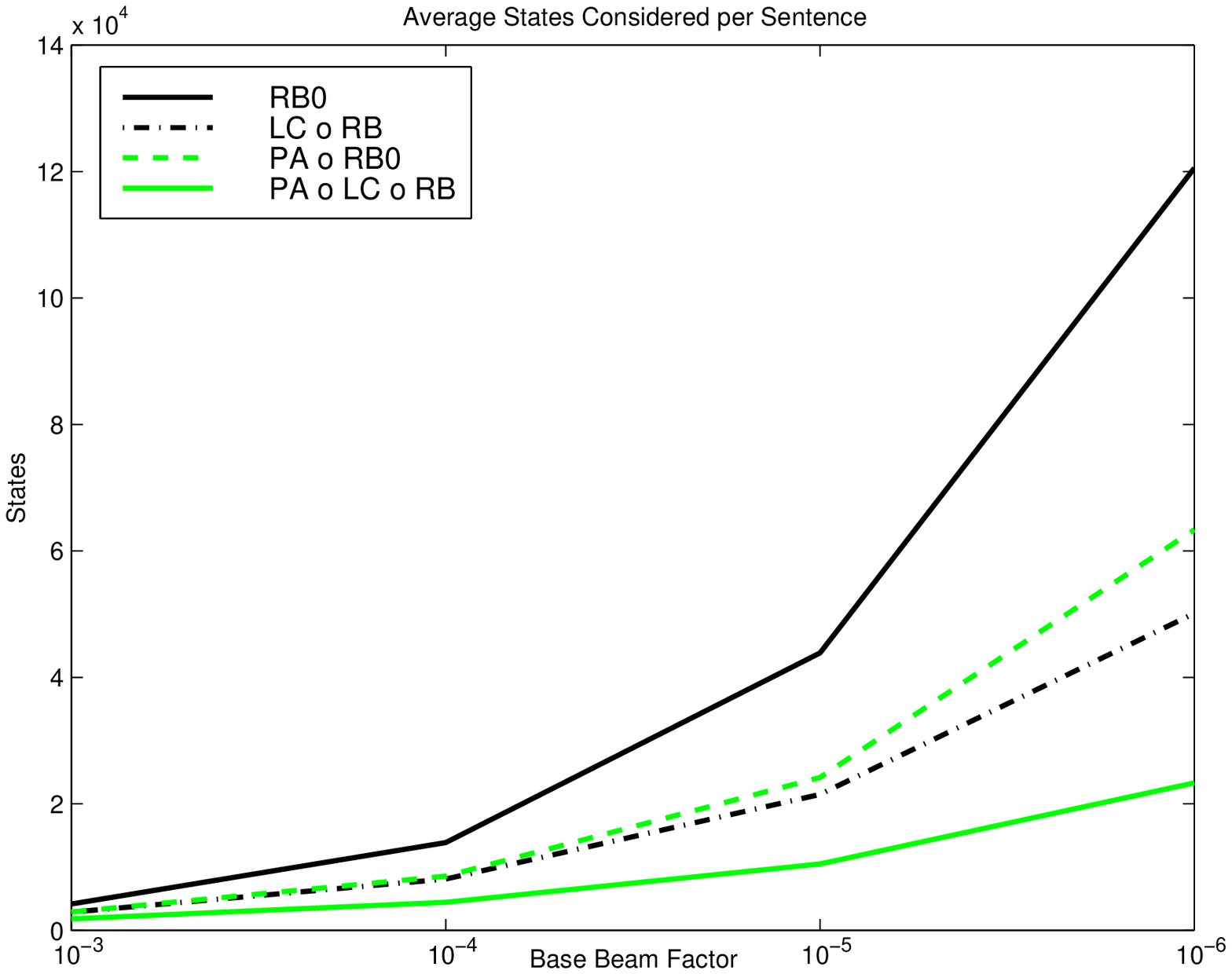, width=4.1in}\vspace*{.05in}\\
\hspace*{1.1in}
\epsfig{file=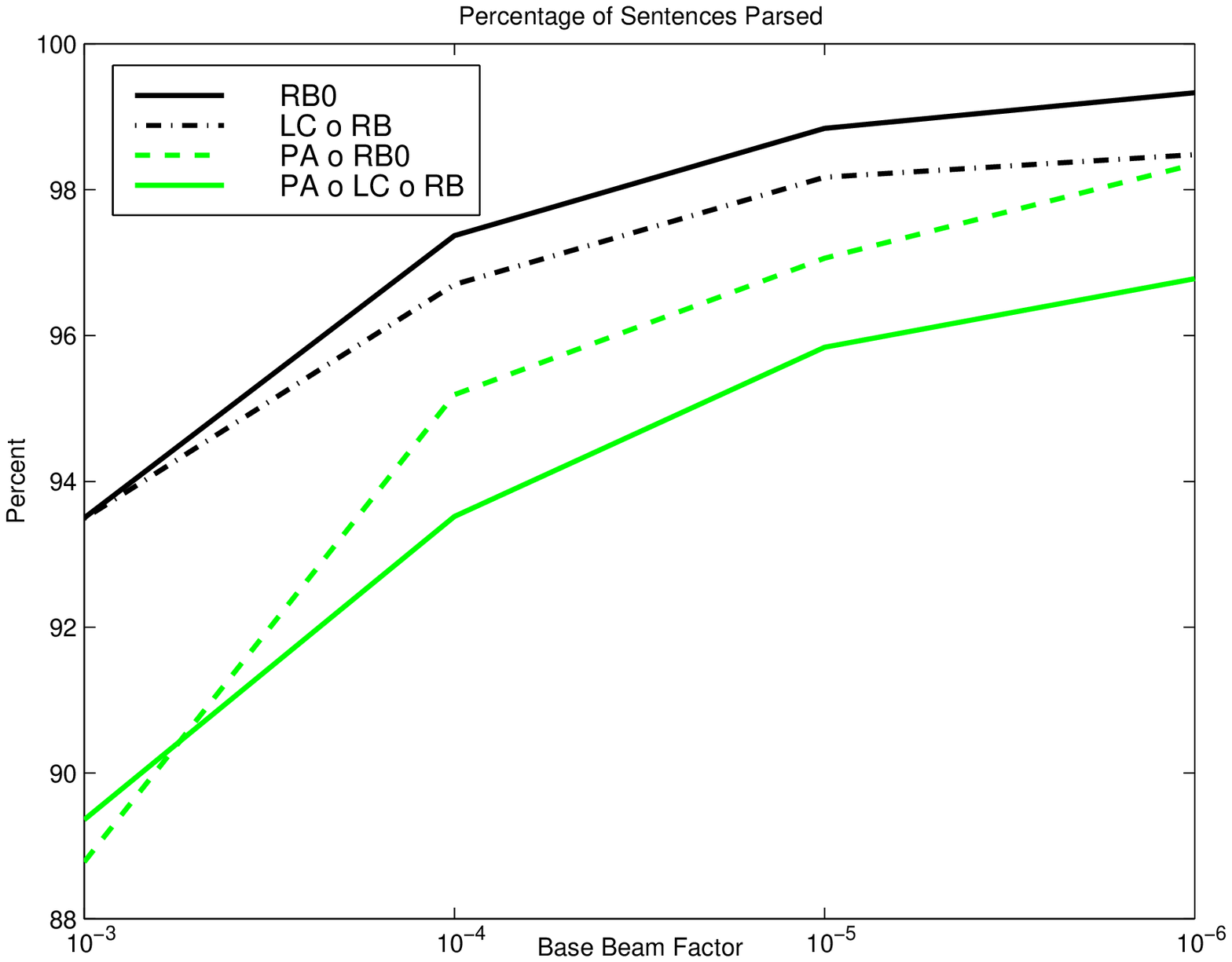, width=4.1in}
\caption{Changes in performance with beam factor variation} \label{fig:ef1}
\end{figure*}

\begin{figure*}
\hspace*{1.1in}
\epsfig{file=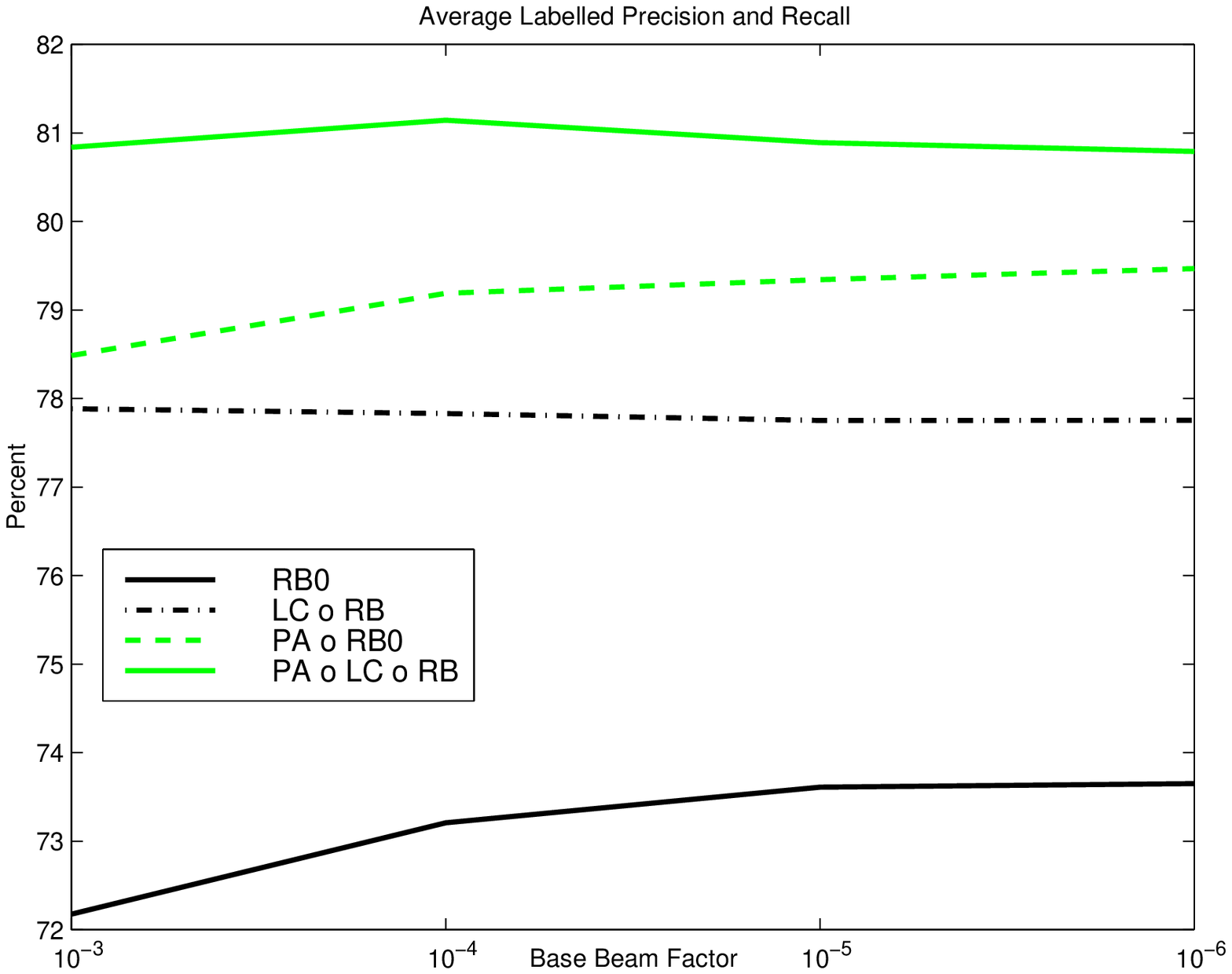, width=4.1in}\vspace*{.05in}\\
\hspace*{1.1in}
\epsfig{file=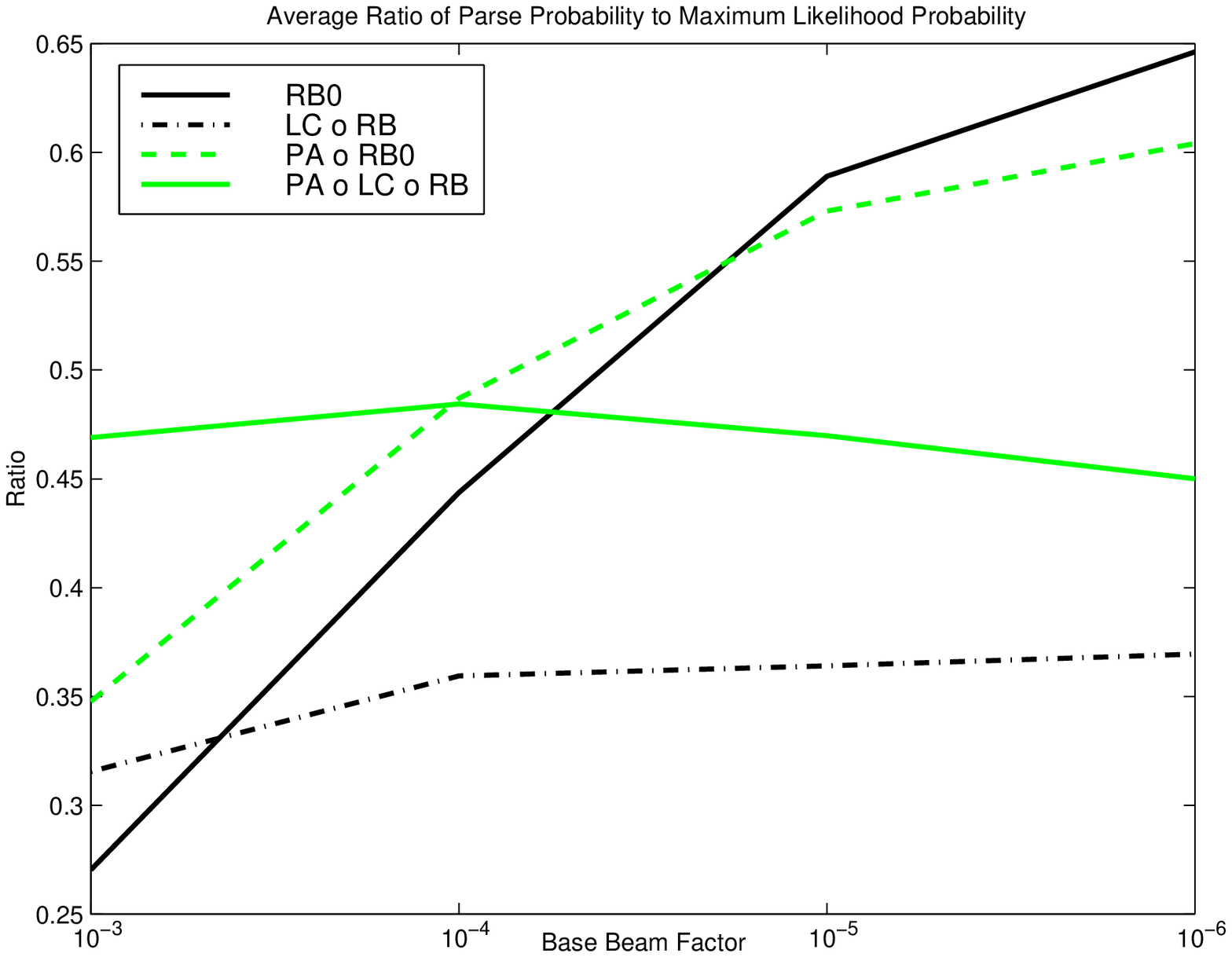, width=4.1in}
\caption{Changes in performance with beam factor variation} \label{fig:ef2}
\end{figure*}

One point that deserves to be made is that there is something of an
accuracy/efficiency tradeoff with regards to the base beam factor. The
results given so far were at $10^{-4}$, which functions pretty well for the
transforms we have investigated. Figures \ref{fig:ef1} and
\ref{fig:ef2} show four performance 
measures for four of our transforms at base beam factors of $10^{-3}$, 
$10^{-4}$, $10^{-5}$, and $10^{-6}$.  There is a dramatically increasing
efficiency burden as 
the beam widens, with varying degrees of payoff. With the top-down
transforms ({\small RB0} and {\small PA} $\circ$ {\small RB0}), the ratio of the average probability
to the {\small MLP} probability does improve substantially as the beam grows,
yet with only marginal improvements in coverage and
accuracy. Increasing the beam seems to do less with the left-corner
transforms. 

\section{Conclusions and Future Research}

We have examined several probabilistic predictive parser variations,
and have shown the approach in general to be a viable one, both in
terms of the quality of the parses, and the efficiency with which they
are found. We have shown that the improvement of the grammars with
non-local information not only results in better parses, but guides
the parser to them much more efficiently, in contrast to dynamic
programming methods. Finally, we have shown that the accuracy
improvement that has been demonstrated with left-corner approaches can
be attributed to the non-local information utilized by the
method. 

This is relevant to the study of the human sentence processing
mechanism insofar as it demonstrates that it is possible to have a
model which makes explicit the syntactic relationships between items
in the input incrementally, while still scaling up to broad-coverage. 

Future research will include:
\begin{list}{$\bullet$}{\setlength{\topsep}{.01in}\setlength{\itemsep}{0in}}
\item lexicalization of the parser
\item utilization of fully
connected trees for additional syntactic and semantic processing
\item the use of syntactic predictions in the beam for language modeling
\item an examination of predictive parsing with a left-branching language
(e.g. German)
\end{list}
In addition, it may be of interest to the psycholinguistic community
if we introduce a time variable into our model, and use it
to compare such competing sentence processing models as race-based
and competition-based parsing.
\bibliography{ber}
\end{document}